\documentclass[10pt,twocolumn,letterpaper]{article}

\usepackage{iccv}
\usepackage{times}
\usepackage{epsfig}
\usepackage{graphicx}
\usepackage{amsmath}
\usepackage{amssymb}
\usepackage{tabularx,ragged2e}
\usepackage{booktabs}
\usepackage{authblk}
\usepackage{soul}

\newcommand{\modelname}{Dynamic Object Generation Network}
\newcommand{\methodname}{neural re-simulation\xspace}

\newcommand{\edit}[1]{#1}

\newcommand{\chapternote}[1]{{%
  \let\thempfn\relax
  \footnotetext[0]{#1}
}}

\renewcommand{\paragraph}[1]{\noindent{\bf #1}}

\usepackage[pagebackref=true,breaklinks=true,colorlinks,bookmarks=false]{hyperref}

\iccvfinalcopy 

\def\iccvPaperID{5023} 

\begin{document}

\title{Neural Re-Simulation for Generating Bounces in Single Images}

\author[1*]{Carlo Innamorati}
\author[2]{Bryan Russell}
\author[2]{Danny M. Kaufman}
\author[1,2]{Niloy J. Mitra}
\affil[1]{University College London}
\affil[2]{Adobe Research}
\affil[ ]{\href{http://geometry.cs.ucl.ac.uk/projects/2019/bounce-neural-resim/}{\texttt{http://geometry.cs.ucl.ac.uk/projects/2019/bounce-neural-resim/}}}


\maketitle

\begin{abstract}
We introduce a method to generate videos of dynamic virtual objects plausibly interacting via collisions with a still image's environment. 
\edit{Given a starting trajectory, physically simulated with the estimated geometry of a single, static input image, we learn to `correct' this trajectory to a visually plausible one via a neural network. The neural network can then be seen as learning to `correct' traditional simulation output, generated with incomplete and imprecise world information, to obtain context-specific, visually plausible re-simulated output -- a process we call neural re-simulation.}
We train our system on a set of 50k synthetic scenes where a virtual moving object (ball) has been physically simulated. We demonstrate our approach on both our synthetic dataset and a collection of 
real-life images depicting everyday scenes, 
obtaining consistent improvement over baseline alternatives throughout. 

\end{abstract}

\section{Introduction}

\begin{quote}
    Christopher Robin: You're next, Tigger. Jump!\\
    Tigger: Er, jump? Tiggers don't jump, they bounce.\\
    Winnie the Pooh: Then bounce down.\\
    Tigger: Don't be ``ridick-orous''. Tiggers only bounce up!\\
    -- A. A. Milne, Winnie The Pooh
\end{quote}

\chapternote{\hspace{-6pt} $\sp*$ \hspace{-4.5pt} Work initiated at Adobe during CI's summer internship.}

A single still image depicts an instant in time. Videos, on the other hand, have the capacity to depict dynamic events where scene objects may interact with and bounce off each other over time. We seek to allow users to bring single still images to life by allowing them to automatically generate videos depicting a virtual object interacting with a depicted scene in the image.

\begin{figure*}[t!]
	\centering
	\includegraphics[width=\textwidth]{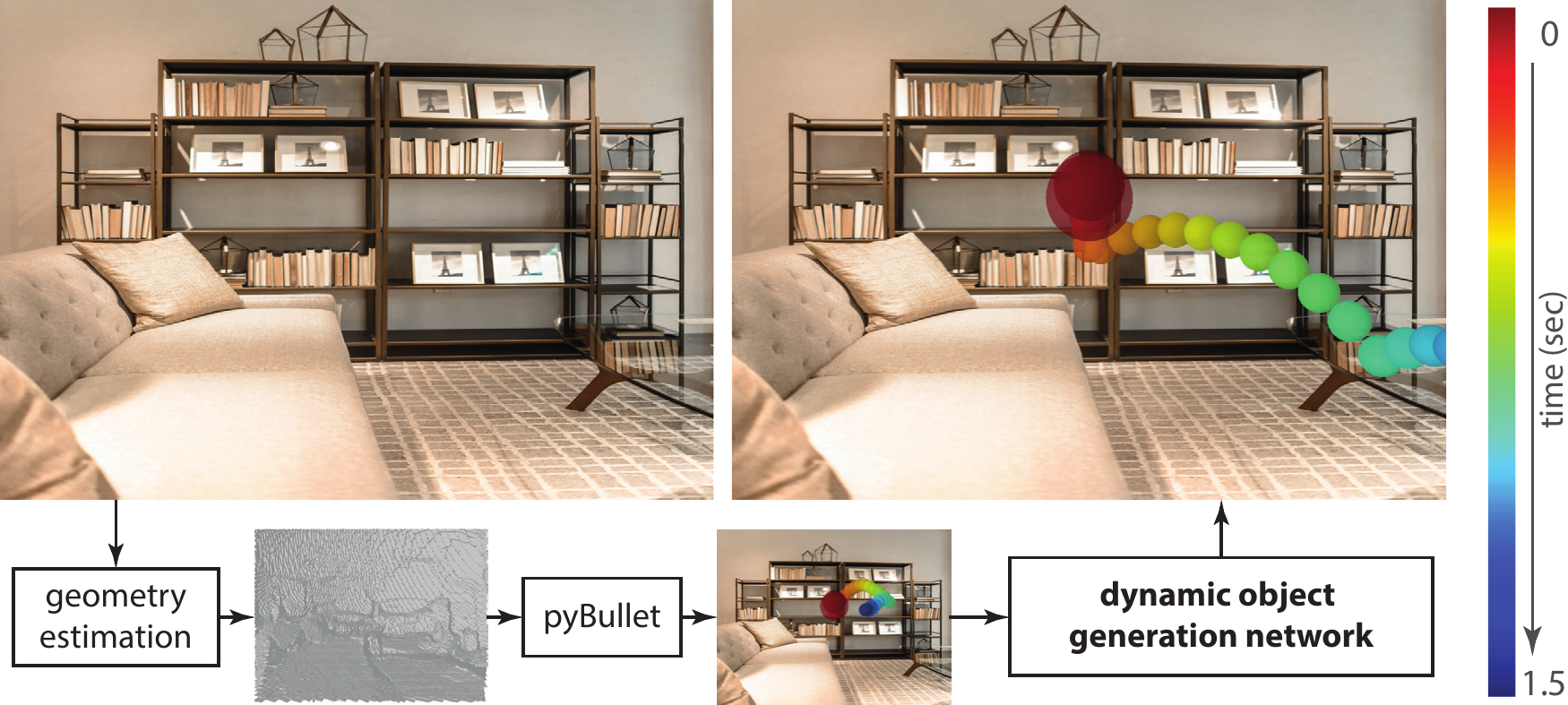}
	\caption{
	{\bf Problem statement and approach overview.} We take as input a single still image depicting a scene and output a video depicting a virtual object dynamically interacting with the scene through bouncing. Here, we consider a ball as our virtual object. We achieve this by our \modelname{} which takes as inputs estimated depth and an initial forward trajectory of the virtual object from the PyBullet simulator~\cite{coumans2017} and outputs a \edit{`corrected' trajectory via a} \methodname step. To visualize all the trajectories in this paper, we composited the virtual object at each time step with the input image; warmer colors indicate earlier time steps. {\bf Please view output videos in the supplemental.} 
	}
	\label{fig:teaser}
\end{figure*}

Specifically, we address the problem of dynamic object compositing in a single still image where a virtual object physically interacts by contact, such as bouncing, with the depicted scene. Our goal is to generate {\em visually plausible} virtual object-scene interactions instead of a 
physically accurate forward prediction. In other words, we seek to generate an output that looks physically valid to a human observer even if it does not exactly match an observed physical interaction starting from the same initial conditions. 
Our task is important for applications in augmented reality and animation, allowing users to author dynamic events in a depicted scene \edit{without access to 
sufficient information about the corresponding world}.

For our study, we focus on the case where we toss a virtual ball into an everyday scene, such as living rooms, bedrooms, and kitchens, and seek to have the ball bounce off different objects in the scene (see Figure~\ref{fig:teaser}). 
This setup presents many challenges as we need to reason about the geometric layout of the scene and infer how the virtual ball will physically interact with the scene. 
Previous approaches either do not physically interact with the scene through contact~\cite{Davis2015ImagespaceMB,Kholgade:2014:OMS},  require captured scene depth through multiple views or active sensing~\cite{Newcombe2011KinectFusionRD,Newcombe2011DTAMDT}, or rely on forward simulation using predicted single-view scene depth. However, predicted scene depth may be noisy, resulting in inaccurate and visually implausible forward-simulated trajectory output (see videos in supplemental). 
Furthermore, there may be global scaling issues with depth predictions if an algorithm is trained on one dataset and applied to another with different camera intrinsics. 

\edit{To address the above challenges, we seek to learn to `re-simulate' the outputs obtained from running a forward simulation over noisy, estimated scene geometry inferred for the input image to yield a visually plausible one.} Re-simulation methods are traditionally applied in visual effects to reuse and combine pre-exisiting physical simulations for new and novel scenarios~\cite{Kim:2013:resimulation,Sato:2018,Thuerey:2016}. Here, in analogy, we introduce a neural network for learning re-simulation -- a process we call {\em neural re-simulation}. In particular, here we apply neural re-simulation from trajectories with noisy and insufficient data to plausible output. Our solution is thus also related to recent approaches for re-rendering scenes with a neural network~\cite{Kim:2018,Martin-Brualla:2018,Meshry19,Thies18}; here we seek to re-simulate dynamic trajectory outputs. 
%
%
%

For the initial forward trajectory, we generate geometry based on estimated depth from the input image and run a physical simulator on this estimated geometry. \edit{Then, in the} \methodname step, our proposed model  `corrects' the initial trajectory resulting from the noisy depth predictions conditioned on context information available in the input image. 
Furthermore, we introduce an approach that learns to correct the global scaling of the estimated depths for the scene conditioned on the initial trajectory. 
We train our system on a dataset of trajectories computed by forward simulating trajectories with an off-the-shelf physical simulator (PyBullet~\cite{coumans2017}) on SUNCG scenes~\cite{song2016ssc}. 
We find that our forward trajectories complement the information provided by the estimated depths. 
Finally, we use an adversarial loss~\cite{GAN_NIPS2014_5423} during training to allow for learning to generate visually plausible trajectory outputs. 

Evaluating the quality of the dynamic object insertion task is difficult due to two factors: first, on real images, there is no available ground truth to compare against; and second, for target applications such as AR/VR and animation, `visual plausibility' is more relevant rather than accurate forward trajectory predictions.  
We evaluate our proposed approach quantitatively both on synthetic data where we have access to ground truth simulations and on real data via a user study. 

Our contribution is a system that learns to correct an initial trajectory of a virtual object provided by forward simulation on geometry specified by predicted depth of an input single still image to output a visually plausible trajectory of the object. Furthermore, we introduce a network that learns to update the predicted depth values conditioned on the initial trajectory. 
We demonstrate our approach on synthetic images from SUNCG and on real images and show that our approach consistently outperforms baseline methods by a healthy margin.

\section{Related Work}
\label{sec:related}

Our work is primarily related to previous approaches on generating video and modeling dynamic object interactions.

\begin{figure*}[t]
	\centering
	\includegraphics[width=\textwidth]{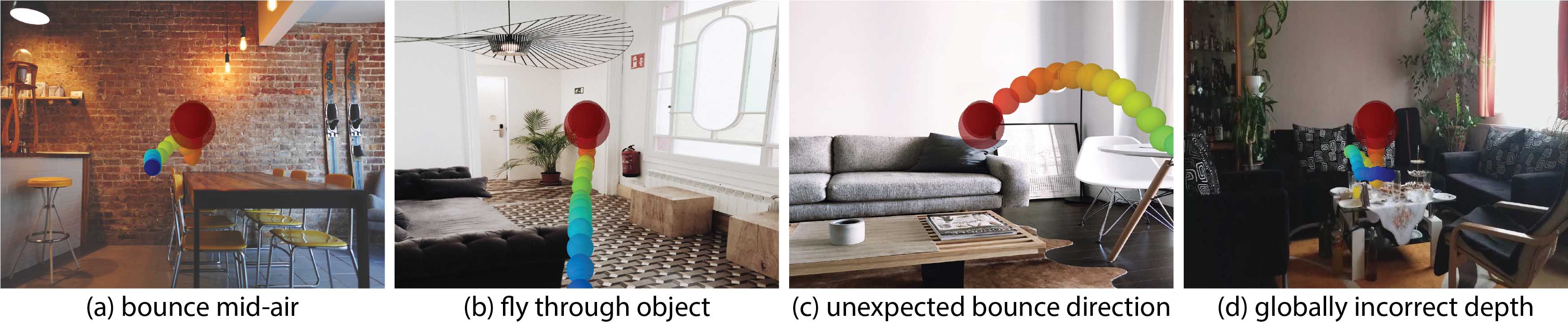}
	\caption{
	{\bf Visually implausible trajectories.} Examples of the visually implausible trajectories that are generated by simulations with depth prediction. Left to right: a virtual object bounces in mid air, flies into an object, bounce in an unexpected direction, or has completely different scale due to globally incorrect depth.
	}
	\label{fig:implausible}
\end{figure*}

\paragraph{Video generation.} 
Prior work has generated dynamic or video textures by analyzing low-level motion features~\cite{Doretto2003DynamicT,Schdl2000VideoT}. However, we seek to model and generate motions due to high-level interactions in a scene. Prior work has looked at modeling interactions in constrained environments, such as sports~\cite{Bettadapura16}. While these interactions are complex and involve reasoning about the intention of multiple agents in a scene, we seek to model interactions that occur in everyday scenes. More recently, there has been work to endow a neural network with the ability to generate video. Examples include forecasting human dynamics from a single image~\cite{chao:2017:cvpr}, forecasting with variational auto-encoders~\cite{vae_eccv2016}, generating visual dynamics~\cite{visualdynamics16}, and generating the future~\cite{vondrick:2017:cvpr}. These works primarily forecast or generate human actions in video and do not focus on modeling object interactions. Moreover, making long-term video generations spanning multiple seconds from a neural network is challenging. Most relevant to our generation task is prior work that disentangles underlying structure and from the generation step~\cite{villegas17hierchvid}.

\paragraph{Object interactions and intuitive physics.}
Prior work in modeling dynamic object interactions have involved reasoning and recovering parameters to a physical system~\cite{Bhat02,brubaker09,Brubaker08,Brubaker10,Kyriazis11,Mann97,Monszpart16,Zhu16,Zhu15}. While these works aim for physical accuracy, our aim is different as we want to achieve visual plausibility through learning. 
Other work has looked at changing the viewpoint of an object in a scene~\cite{Kholgade:2014:OMS} or manipulating modal bases of an object~\cite{Davis2015ImagespaceMB}, but do not address the problem of object-scene interaction via contact. 
Recent work has aimed to train a learning system to reason about physics for understanding the semantics of a scene~\cite{gupta_eccv10} or for making future predictions in synthetic 2D scenarios~\cite{BattagliaNIPS2016,Chang17,Fragkiadaki2016,Watters:2017:VIN}. Work has aimed to go beyond the synthetic setting by inferring forces in real-world images through Newtonian image understanding~\cite{Mottaghi16b}. More recently, work has aimed to learn to model 3D systems, such as predicting where toy blocks in a tower will fall~\cite{Lerer:2016}, modeling a variety of different closed-world systems such as ramps, springs and bounces~\cite{EhrhardtMMV17,EhrhardtMVM17,Wu16,Wu15}, and predicting the effect of forces in images by reasoning over the 3D scene layout and training over 3D scene CAD models~\cite{Mottaghi16}. Recent work has also leveraged learning about real-world interactions by interacting with the world through visual-motor policies~\cite{Levine:2016}, grasping~\cite{Pinto:Gupta:2016}, pushing and poking objects~\cite{Agrawal:2016:nips,Pinto:Gupta:2016b}, crashing a drone into scene surfaces~\cite{Gandhi17}, hitting surfaces with a stick to reason about sound~\cite{Owens16}, or generating audible shapes~\cite{Zhang17}. Closest to ours is recent work that uses a neural network to make forward predictions of bounces in real-world scenes~\cite{Purushwalkam19} and infer scene geometry and physical parameters such as coefficient of restitution~\cite{Wang:2017:BounceMaps}. Note that this work, while similar to ours, aims for physical accuracy and predicts an immediate short-term trajectory of a ball bounce after making contact with a surface. We go beyond this work and infer roll outs over multiple bounces and aim for visual plausibility. 

\section{\modelname{}}
\label{sec:approach}

\newcommand{\image}{\mathcal{I}}
\newcommand{\initialconditions}{\rho}
\newcommand{\video}{\mathcal{V}}
\newcommand{\depth}{Z} 
\newcommand{\initialdepth}{\depth_0}
\newcommand{\trajectory}{X}
\newcommand{\trajectorystate}{x}
\newcommand{\noise}{z}
\newcommand{\paramdepth}{\theta}
\newcommand{\updatetrajectory}{\mathcal{G}_\noise}
\newcommand{\updatetrajectorynonoise}{\mathcal{G}}
\newcommand{\updatedepth}{\mathcal{H}}
\newcommand{\timelength}{T}
\newcommand{\initialtrajectory}{\trajectory_0}
\newcommand{\simulator}{\mathcal{S}}
\newcommand{\compose}{\mathcal{R}}
\newcommand{\discriminator}{\mathcal{D}}
\newcommand{\dataplausible}{p_{plausible}}
\newcommand{\dataimplausible}{p_{initial}}
\newcommand{\gaussian}{\mathcal{N}{\left(0,1\right)}}

\begin{figure*}[t]
	\centering
	\includegraphics[width=\textwidth]{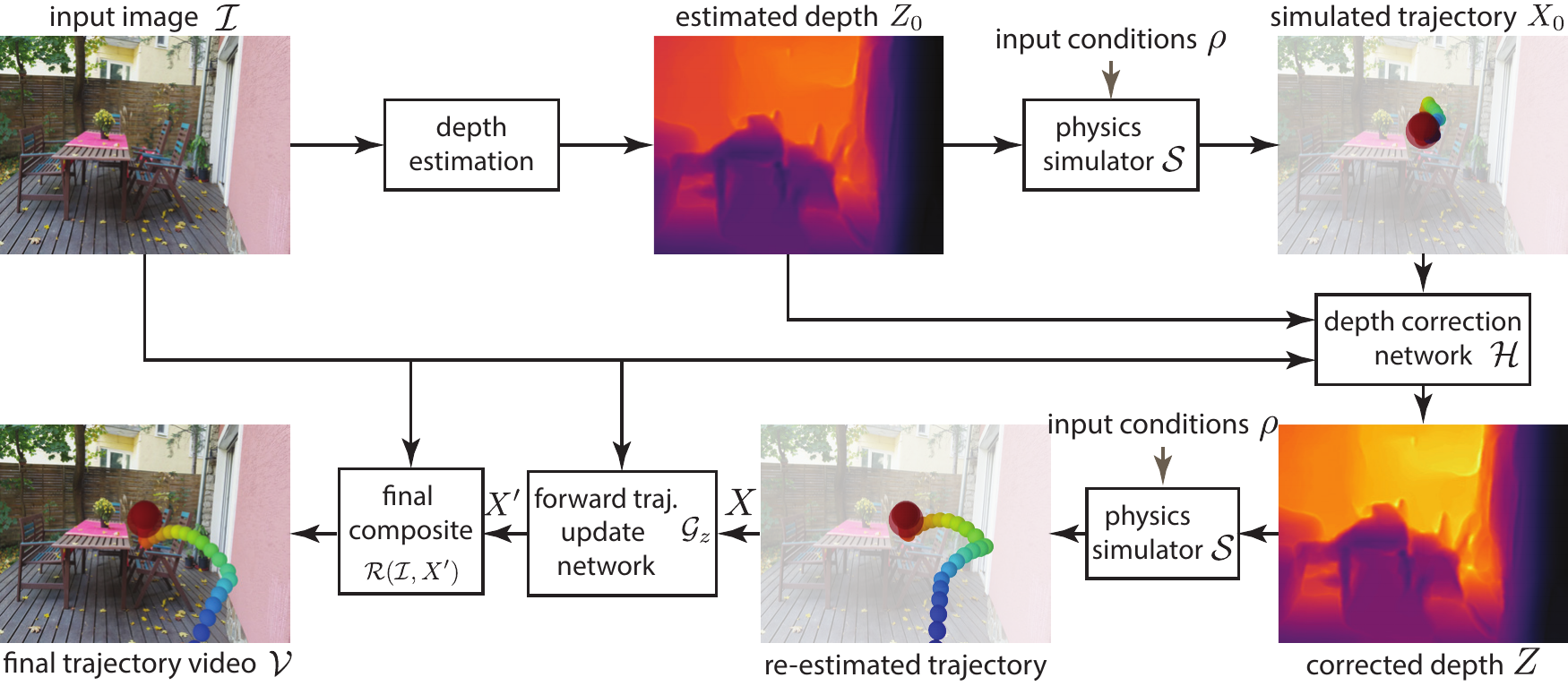}
	\caption{
	{\bf System overview.} Our system takes as inputs an image depicting a scene and initial conditions for the object that is tossed in the scene and outputs a video showing a visually plausible predicted trajectory of the object interacting with the scene. Our approach predicts depth at every pixel in the image and consists of two networks -- a forward trajectory update network $\updatetrajectory$ and a depth correction network $\updatedepth$. See text for more details. 
	}
	\label{fig:model}
\end{figure*}

Given an input image $\image$ depicting a scene, we seek to output a video $\video$ of time length $\timelength$ where a virtual moving object with initial conditions $\initialconditions$ has been composited into the image depicting physical interaction with the scene. Example scene interactions include the virtual object flying in the air, making contact with several scene objects, and changing its trajectory after bouncing off scene objects. While one could learn to generate the video directly from training data using a neural network, such long-term generations are currently hard due to fundamental issues such as temporal flickering and decaying visual signal to the mean~\cite{Mathieu2016}. Recent work in long-term video generation has shown success by disentangling the prediction of the underlying scene structure and generation of the video pixels~\cite{villegas17hierchvid}. We seek to leverage this insight for our video generation task. 

Our approach starts by computing an initial depth map $\initialdepth$ of the input image, which is then passed, after geometry processing, to a physical simulator $\simulator$ along with initial conditions $\initialconditions$. The initial conditions consist of the initial state of the virtual object (velocity and position) and material parameters (\eg, coefficients for friction and restitution).
The output of the physical simulator is an initial trajectory $\initialtrajectory=\simulator{\left(\initialdepth,\initialconditions\right)}$ represented as a sequence of object states $\initialtrajectory = \left(\trajectorystate_1,\dots,\trajectorystate_\timelength\right)$. 
We assume access to depth predictions for the scene from an off-the-shelf algorithm. For our work, we use the depth prediction system of Chakrabarti \etal~\cite{Chakrabarti2016DepthFA}, which is a top performer on predicting depth on the NYUv2 dataset~\cite{Silberman2012IndoorSA}. 
We run forward simulations using the PyBullet physical simulator~\cite{coumans2017}.

Given the initial forward trajectory $\initialtrajectory$, one could simply generate the final video $\video$ by passing the trajectory and the image $\image$ to a composing function $\compose$ to yield $\video=\compose{\left(\image,\initialtrajectory\right)}$. However, this step often yields visually implausible output videos where the virtual object may or may not change direction at inappropriate times or bounce in a direction that is not congruent with a depicted surface due to inaccurate depth predictions. Example visually implausible outputs are shown in Figure~\ref{fig:implausible}. Such visually implausible artifacts cause the viewer to perceive the virtual object to bounce in mid-air, fly into a scene object or surface, or bounce in an unexpected direction, and are due to inaccurate predictions in the depth estimates. 

Our insight is to correct such visually implausible artifacts by correcting the initial depth $\initialdepth$ and initial forward trajectory $\initialtrajectory$ to yield a visually plausible final result -- \edit{we call this step} \methodname{}. We introduce two networks $\updatetrajectory$ and $\updatedepth$ that generate updated forward trajectory $\trajectory$ and updated depth map $\depth$, respectively. 
The forward trajectory update network $\updatetrajectory$ is a generative neural network parameterized by scalar $\noise$ that takes as inputs the image $\image$ and a forward trajectory $\trajectory$ and returns an updated trajectory, 
\begin{equation}
    \trajectory^\prime = \updatetrajectory{\left(\image,\trajectory\right)}.
\end{equation}
The depth correction network $\updatedepth$ is a neural network that returns an updated depth map given the image $\image$, initial depth map $\initialdepth$, and initial trajectory $\initialtrajectory$ as inputs, 
\begin{equation}
    \depth = \updatedepth{\left(\image,\initialdepth,\initialtrajectory\right)}.
\end{equation}

The final video can be generated by composing the two networks,
\begin{equation}
    \video = \compose{\left(\image,\updatetrajectory{\left(\image,\simulator{\left(\updatedepth{\left(\image,\initialdepth,\initialtrajectory\right)},\initialconditions\right)}\right)}\right)}.
\end{equation}
We describe both networks in the following subsections and outline our overall approach in Figure~\ref{fig:model}.

\subsection{Trajectory update network}
\label{sec:update_network}

We assume a neural network for the trajectory update network. 
The network takes as inputs the input image $\image$, the virtual object's forward trajectory $\trajectory$ resulting from the corrected depth predictions for the scene, and a value $z$ sampled from a Gaussian distribution $\noise\sim\gaussian$. 
The network first consists of a multilayer perceptron (MLP) that takes as input a concatenation of the trajectory $X$ and sampled value $z$ and outputs an encoded representation of the trajectory. The MLP is followed by a second MLP that takes as input a concatenation of the encoded trajectory representation and an encoding of the image $\image$. We obtain the image encoding through a pre-trained Inception-ResNet model~\cite{SzegedyIVA17} that has been pre-trained on ImageNet. The encoding is extracted from the penultimate layer. The second MLP outputs the updated trajectory $\trajectory^\prime$. 

\paragraph{Learning.} 
One could train our trajectory update network using an $L_2$ loss to ground truth trajectories with the aim of making physically accurate predictions. However, this strategy would yield over-smooth predictions to the mean distribution over trajectories. Moreover, our goal is to generate visually plausible trajectories and not necessarily physically accurate ones.

To achieve our goal, we aim to fool a discriminator given a dataset of trajectories and the set of initial trajectories provided by running a forward simulation on the geometry from the initial depth map $\initialdepth$. We train the trajectory update network using an adversarial loss~\cite{GAN_NIPS2014_5423}. Given training examples of visually plausible trajectories $\dataplausible$ and a set of initial trajectories $\dataimplausible$, we seek to optimize the following adversarial loss over the trajectory update network $\updatetrajectorynonoise$ and a discriminator architecture $\discriminator$,

\begin{align}
    \min_{\updatetrajectorynonoise} \max_{\discriminator} \ & \mathbb{E}_{\trajectory\sim \dataplausible}{\left[\log{\left(\discriminator{\left(\image,\trajectory\right)}\right)}\right]} \nonumber \\ 
    & + \mathbb{E}_{\substack{\trajectory\sim\dataimplausible \\
    \noise\sim\gaussian}}{\left[\log{\left(1-\discriminator{\left(\image,\updatetrajectory{\left(\image,\trajectory\right)}\right)}\right)}\right]}.
\end{align}

The discriminator network consists of an MLP that takes as input a trajectory $X$ and outputs its encoded representation. The MLP is followed by a second MLP that takes as input a concatenation of the encoded trajectory representation and an Inception-ResNet encoding of the image $\image$. The second MLP outputs a prediction label.

To help with the early stages of training, the adversarial loss is aided by an L2 loss of decreasing relevance over training epochs. In particular, the L2 loss is weighted down by 0.5\% after every epoch, resulting in a complete adversarial loss after the 200th epoch. The network is run for a total 1k epochs.
\begin{figure}[t!]
	\includegraphics[width=\linewidth]{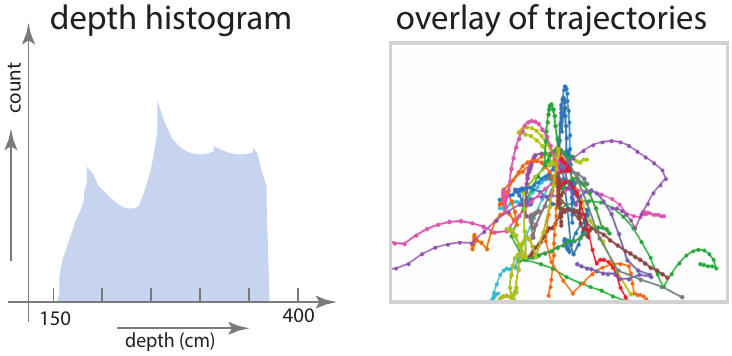}
	\caption{
	{\bf Dataset.} Histogram of scene depth (left) and sampled trajectories from our dataset (right), illustrating the dataset's variety over depth and trajectory.
	}
	\label{fig:dataset}
\end{figure}

\subsection{Depth correction network}

A major source of error is when an initial depth map $\initialdepth$ is grossly out of range of the expected depth values for a given depicted scene. To correct this issue, we seek to have a network learn to output calibration parameters $\depth_{min}$ and $\depth_{max}$ that will be used to scale the initial depth values into the expected range. To achieve this result, we assume the depth correction network $\updatedepth$ is a MLP followed by a depth calibration update that takes as input a concatenated vector consisting of an encoded representation of the input image $\image$, an encoding of the initial depth map $\initialdepth$, and the output of the trajectory update network with the initial trajectory $\initialtrajectory$ passed as input. The MLP outputs $\left(\depth_{min},\depth_{max}\right)$, which is then used to scale the min and max input depth map values in $\initialdepth$ to match the newly computed normalization values. The updated depth map $\depth$ is returned as output. 

\paragraph{Learning.} 
The depth correction network is trained using $L_2$ loss to regress the two normalization parameters given the trajectory $\initialtrajectory$, an encoding of the image $\image$ and an encoding of the initial depth map $\initialdepth$. The network is run for a total of 1k epochs.

\paragraph{Geometry processing.} 
To obtain the input trajectories of the model $X_0$ from the input depth map, we leverage projection and view matrices to obtain a point cloud, where each vertex corresponds to a pixel of the depth map. We then turn the point cloud into a mesh by connecting neighboring vertices. The mesh is then passed through PyBullet to obtain a corresponding trajectory. 



\section{Synthetic Trajectory Generation}
\label{sec:dataset_synthetic}

A critical aspect of training our system is the ability to learn from a large collection of examples depicting an object interacting with an everyday scene. This aspect is challenging as such data is relatively scarce. For example, while one could consider real videos, the largest known dataset of a ball interacting with a scene contains about 5000 videos~\cite{Purushwalkam19}. Moreover, the ball starts with different initial velocity (speed and direction) in each video. 

To overcome this challenge, we leverage recent datasets containing large stores of 3D CAD models. We consider the SUNCG dataset for our study~\cite{song2016ssc}. The SUNCG dataset contains 45k 3D scenes. We equip the 3D CAD models with a physics simulator, namely PyBullet~\cite{coumans2017}. To render forward trajectories, we import a scene into PyBullet and specify a camera viewpoint. We leverage the pre-computed camera locations provided by the SUNCG toolbox, which depicts viewpoints where the camera is held upright with the pitch angle rotated up by 30 degrees and pointing toward the interior of the scene geometry. We filtered the cameras to not directly face walls and other large surfaces and adjusted the cameras to match PyBullet's intrinsic parameters. We use 50k of the available 828k cameras. 

For our study, we assume that the object is a spherical ball and starts with an initial velocity of 0.6 meters per second in the direction away from the camera center. 
We set the coefficients for friction and restitution to 0.5 and generate forward trajectories of length 1.5 seconds, sampled at 20Hz. At each time point, we record the output of the 3D center-of-mass location of the ball in camera space from PyBullet. After processing the trajectory, we render the new frames by re-creating the object in PyBullet and re-updating the coordinate system.

Our dataset consists of 50k examples, with each example coming from a multi-room 3D model containing on average eight different rooms. 
We generate data using less than 10\% of the available cameras; the majority of the rooms have less than two sampled viewpoints out of maximum five.
We split our dataset into training, validation, and testing sets of sizes 40k, 5k and 5k, respectively. 
Figure~\ref{fig:dataset} shows a histogram of the span of depth values over the dataset in centimeters, in addition to a set of randomly selected trajectories from the dataset, illustrating the dataset's variety.


\begin{figure*}[t!]
	\centering
	\includegraphics[width=\textwidth]{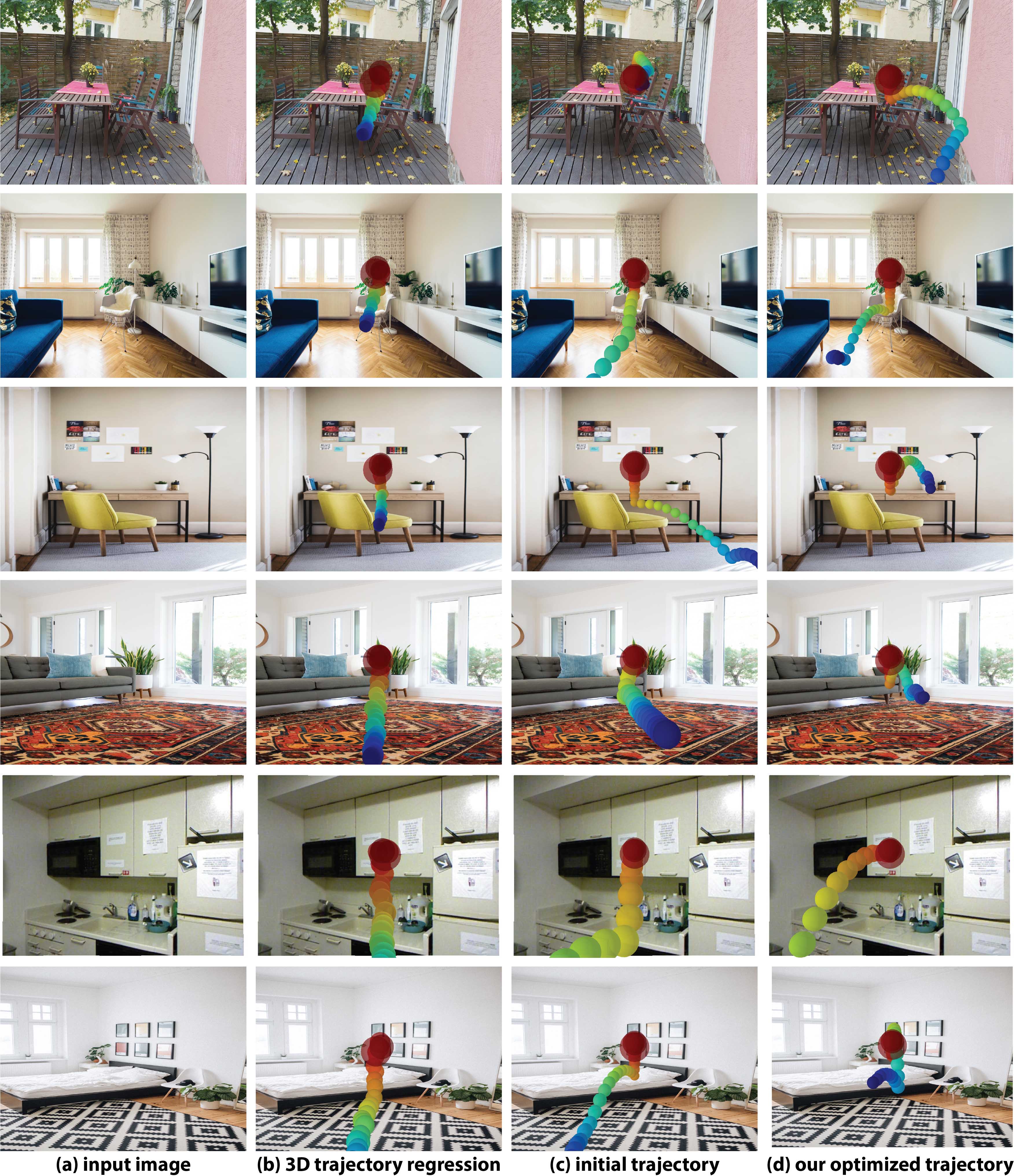}
	\caption{
	{\bf Qualitative results on real images.} 
	We show (a) the input image depicting a real scene, (b) output from the 3D trajectory regression baseline, (c) our initial trajectory resulting from forward simulation on predicted depth, and (d) our output optimized trajectory. Notice how the visual plausibility of our output trajectories improve over the initial trajectory. Last row -- failure example. 
	}
	\label{fig:qualitative_real}
\end{figure*}

\section{Experiments}

\begin{table}[t]
\caption{\label{tab:quantitative} 
{\bf Quantitative evaluation on synthetic scenes.} We report $L_2$ distances in 2D and 3D and a perceptual loss (top -- baselines, bottom -- ablations). Notice how our approach out-performs all baselines and ablations across all criteria.
}
\vspace{.1in}
\centering 
\resizebox{\linewidth}{!}{%
\begin{tabular}{lrrr} 
\toprule
Method & $L_2$ (2D) & $L_2$ (3D) & Perc.\ loss\\
& $\left(1\times10^4\right)$ & $\left(1\times10^4\right)$ & $\left(1\times10^2\right)$\\
\midrule

Dataset prior &  616.4 & 797.8 & 6.3 \\
Depth + fwdS. &  186.5 & 255.1 & 4.9 \\
DepthEq + fwdS. &  144.1 & 196.3 & 4.5 \\
2D regression &  4.9 & N/A & N/A \\
3D regression &  4.9 & 6.5 & 3.1 \\
\midrule
DepNet &  101.2 & 133.5 & 2.9 \\
TrajNet & 3.3 & 4.3 & 2.4 \\
\textbf{Ours}  &  \textbf{1.5} & \textbf{2.1} & \textbf{1.3} \\




\bottomrule
\end{tabular}
} 
\end{table}

In this section, we show qualitative and quantitative results of our system. To better appreciate our final results, we encourage the reader to view videos of our dynamic composite outputs in the supplemental material.

\subsection{Results on synthetic data}
\label{sec:results_synthetic}

As a first experiment, we evaluate the effectiveness of our approach on synthetic scenes from the SUNCG dataset~\cite{song2016ssc}, which allows us to directly compare against trajectories resulting from forward simulation. 

\paragraph{Dataset and evaluation criteria.} 
We use the generated trajectories resulting from forward simulation as outlined in Section~\ref{sec:dataset_synthetic}. 
We evaluate our predicted trajectories by comparing against ground truth trajectories using $L_2$ distance averaged over time. While this criterion evaluates physical accuracy of the predicted trajectory, it does not evaluate the trajectory's visual plausibility. In addition to reporting time-averaged $L_2$ distance, we also report a perceptual loss over the rendered video by computing the $L_2$ distance between the image encoding of each frame with the corresponding ground truth frame. We obtain the image encodings through a pre-trained Inception-ResNet model that has been pre-trained on ImageNet~\cite{SzegedyIVA17} and compare the responses from the penultimate layer.


\paragraph{Baselines.} 
We evaluate a number of baselines for our task. First, we consider a baseline ({\em Dataset prior}) where we compute the average 3D forward trajectory over the training set. Second, we train a neural network to regress to 2D and 3D trajectories ({\em 2D regression} and {\em 3D regression}) given the input single image. For fair comparison, the networks share the same architecture as the trajectory update network outlined in Section~\ref{sec:update_network} except the sampled value $z$ is withheld as input; for 2D, we used only two dimensions for input/output. We trained the networks for the same number of epochs while monitoring the validation loss to avoid over-fitting. 
Third, we consider a baseline of a forward physical simulation of the ball using geometry from the predicted depth ({\em Depth + fwdS.}). We used the same depth-prediction algorithm~\cite{Chakrabarti2016DepthFA} as in our proposed method. Finally, we consider performing histogram equalization over histogrammed ground truth and predicted depth values over the training set ({\em DepthEq + fwdS.}).

\paragraph{Ablations.} 
We consider the following ablations of our model. First, we consider running our full pipeline without the depth correction network ({\em TrajNet}). Second, we consider running our full pipeline without the last trajectory update network ({\em DepNet}). 

\paragraph{Results.} 
To evaluate the improvement we achieve with our depth correction network, in addition to standard $L_2$ measures, we computed the difference in time to the first bounce event (lower is better): Dataset prior -- 3.8, Depth + fwdS.\ -- 3.0, DepthEq + fwdS.\ -- 2.6, DepNet -- 1.8; our depth correction network outperforms the baselines on this criterion.
We show final quantitative results in Table~\ref{tab:quantitative}. Notice how we outperform all baselines and ablations across all criteria.

\subsection{Results on real data}
\label{sec:results_real}

As a second experiment, we evaluate the effectiveness of our approach on single still images depicting real scenes. As per the synthetic experiments, we trained our system on the synthetic data from Section 4.

\paragraph{Dataset, baselines, ablations.} 
We collected a dataset of 30 {\em in the wild} natural images depicting indoor scenes from royalty-free sources and compared our approach against the baselines and ablations described in Section~\ref{sec:results_synthetic}. Additionally, we randomly selected a set of 20 NYUv2 images~\cite{Silberman2012IndoorSA} and compared our approach against forward simulation on the provided depths from an active-sensing camera. Lastly, we compared our method to the work described in~\cite{Purushwalkam19}.

    
\paragraph{Qualitative results.} 
We show qualitative results in Figure~\ref{fig:qualitative_real}. Notice how our approach generally improves over the initial trajectories and out-performs the 3D trajectory regression baseline that returns trajectories close to the mean. 

\paragraph{User study.} 
As we do not have noise-free ground truth trajectories for either sets of images, we conducted a user study where we asked humans to judge the visual plausibility of the outputs. 
More specifically, we presented a user with two outputs from different systems and asked the user to choose which output looks more realistic. 
We randomized the order in which we showed each output to the user. 
The experiments were conducted with workers from thehive.ai. 

For the set of 30 natural images, we compared our results against forward simulation on predicted depth (Depth + fwdS.) and our full pipeline without the last trajectory update network (DepNet).  Additionally, we also compared Depth + fwdS. against DepNet. Each experiment was conducted by 80 unique users and the users casted 4.5k votes over the three tasks. 
Users preferred our method over Depth + fwdS.\ and DepNet 71\% and 59\% of the time ($p < .0001$ -- all p-values from binomial test), respectively, illustrating the effectiveness of our approach. 
Moreover, users preferred DepNet over Depth + fwdS.\ in 63\% of the cases ($p < .0001$), illustrating that the depth correction network helps improve results. 
For the NYUv2 images, we evaluated our method against forward simulation on the provided active-sensing depths over 98 unique users casting 2k votes. Users preferred our method 49\% of the time (no statistical significance), demonstrating the effectiveness of our approach and the level of noise in the active-sensing depths. 

\begin{figure}[t!]
\begin{center}
   \includegraphics[width=\linewidth]{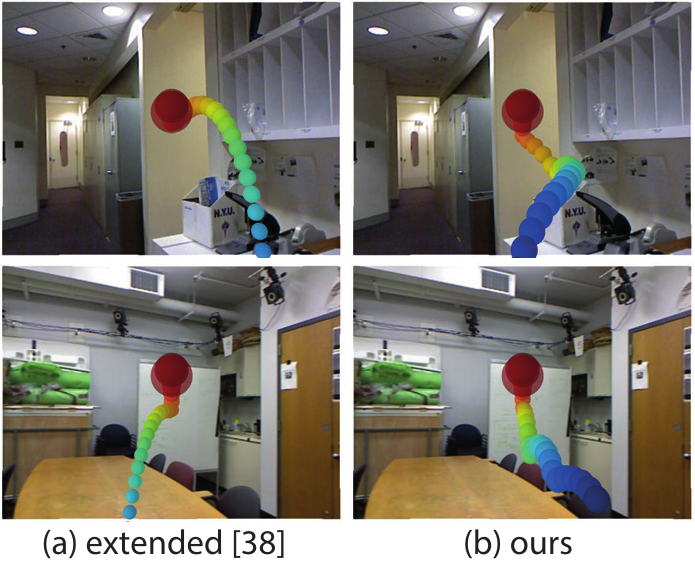}
\end{center}
   \caption{\textbf{Sample user study trajectories.}  We show
   our results versus results obtained by providing access to ground-truth depth and extending the work of~\cite{Purushwalkam19} through free fall. Note that, in ground-truth augmented~\cite{Purushwalkam19} (see text), the ball passes through scene objects, such as the cubicle (row 1) and table (row 2).}
\label{fig:comparison}
\end{figure}

Finally, we compared to the work of~\cite{Purushwalkam19}. As a direct comparison with~\cite{Purushwalkam19} is not feasible due to the additional information it requires, we designed a \textit{ground-truth-augmented version} of the method. The work in ~\cite{Purushwalkam19} requires a ground-truth input trajectory up to the first bounce, which we can provide by running PyBullet with appropriate parameters over the geometry obtained from the kinect depths of the NYUv2 dataset. Further, as~\cite{Purushwalkam19} outputs a post-bounce trajectory spanning only 0.1s and our approach outputs multiple bounces and roll-out trajectories over 1.5s, we manually extended the output from ~\cite{Purushwalkam19} to 1.5s via 3D parabola fitting (note that this step is comparable to simulating post-bounce free fall). We show a qualitative comparison of the resulting trajectories in Figure~\ref{fig:comparison}.

For the experiment conducted on the aforementioned data, 130 users cast a total of 2.5k votes. Users preferred our method in 59\% of the cases $(p < .0001)$, which is noteworthy given that~\cite{Purushwalkam19} has access to ground-truth about when and where in the scene the bounce occurs and indirect access to kinect depths for the scene. Additionally, we performed a second study, where our trajectories were extended after the first bounce through free fall -- note that this is an ablated version of our method as it lacks multi-bounces and roll-outs. For this study, 90 users cast a total of 1.8k votes. Users preferred our outputs 56\% of the time $(p < .0001)$.

\section{Conclusion}

We \edit{introduced} \methodname as a `correction' mechanism that learns to generate visually plausible bounce interactions of a virtual ball in depicted scenes in single still images. Our system learns to update an initial depth estimate of the depicted scene through our depth correction network and uses this update to correct an initial trajectory obtained via forward simulation through our trajectory update network. We demonstrated our system on not only synthetic scenes from SUNCG, but also on real images. We showed via a human study that our approach on real images yields outputs that are more visually plausible than baselines. Our approach opens up the possibility of generating more complex interactions in single still images, such as inserting objects with different geometry and physical properties and modifying the depicted environment. 

\section*{Acknowledgements}

We thank our reviewers for their insightful comments. We also thank Senthil Purushwalkam for helping out with the comparison with~\cite{Purushwalkam19} and Tobias Ritschel, Paul Guerrero and Yu-Shiang Wong for their technical help throughout the project. This work was partially funded by the ERC Starting Grant SmartGeometry (StG-2013-335373).

{\small
\bibliographystyle{ieee_fullname}
\bibliography{main}
}

\end{document}